\documentclass[10pt,twocolumn,letterpaper]{article}

\usepackage{cvpr}
\usepackage{times}
\usepackage{epsfig}
\usepackage{graphicx}
\usepackage{amsmath}
\usepackage{amssymb}
\usepackage{subfigure}
\usepackage{bbm}
\usepackage{amssymb}
\usepackage{symmath}
\usepackage{algorithm}
\usepackage{algorithmic}
\DeclareMathOperator*{\argmin}{arg\,min}

\usepackage[breaklinks=true,bookmarks=false,colorlinks]{hyperref}

\cvprfinalcopy %

\ifcvprfinal\fi
\begin{document}

\title{Binarizing MobileNet via Evolution-based Searching}
\author{\hspace{-0.8cm} Hai Phan$^{1}$ \thanks{indicates equal contribution.} \hspace{0.3cm} Zechun Liu$^{1,3}$ $^*$ \hspace{0.3cm} Dang Huynh$^2$ \hspace{0.3cm}  Marios Savvides$^1$ \hspace{0.3cm}Kwang-Ting Cheng$^3$ \hspace{0.3cm}  Zhiqiang Shen$^1$\\
\hspace{-0.5cm}$^1$Carnegie Mellon University \hspace{0.2cm} $^2$Axon Enterprise \hspace{0.2cm} $^3$Hong Kong University of Science and Technology\\
{\tt \small \{haithanp,marioss,zhiqians\}@andrew.cmu.edu \hspace{0.5cm} dhuynh@axon.com \hspace{0.5cm} \{zliubq,timcheng\}@ust.hk \hspace{1cm}}
}

\maketitle
\thispagestyle{empty}

\begin{abstract}
Binary Neural Networks (BNNs), known to be one among the effectively compact network architectures, have achieved great outcomes in the visual tasks. Designing efficient binary architectures is not trivial due to the binary nature of the network. In this paper, we propose a use of evolutionary search to facilitate the construction and training scheme when binarizing MobileNet, a compact network with separable depth-wise convolution. Inspired by one-shot architecture search frameworks, we manipulate the idea of group convolution to design efficient 1-Bit Convolutional Neural Networks (CNNs), assuming an approximately optimal trade-off between computational cost and model accuracy. Our objective is to come up with a tiny yet efficient binary neural architecture by exploring the best candidates of the group convolution while optimizing the model performance in terms of complexity and latency. The approach is threefold. First, we train 
strong baseline binary networks with a wide range of random group combinations at each convolutional layer. This set-up gives the binary neural networks a capability of preserving essential information through layers. Second, to find a good set of  hyper-parameters for group convolutions we make use of the evolutionary search which leverages the exploration of efficient 1-bit models. Lastly, these binary models are trained from scratch in a usual manner to achieve the final binary model.
Various experiments on ImageNet are conducted to show that following our construction guideline, the final model achieves \textbf{60.09\%} Top-1 accuracy and outperforms the state-of-the-art CI-BCNN with the same computational cost.
\end{abstract}

\section{Introduction}
\label{sec:intro}
\begin{figure}[h!]
    \centering
    \includegraphics[scale=0.4]{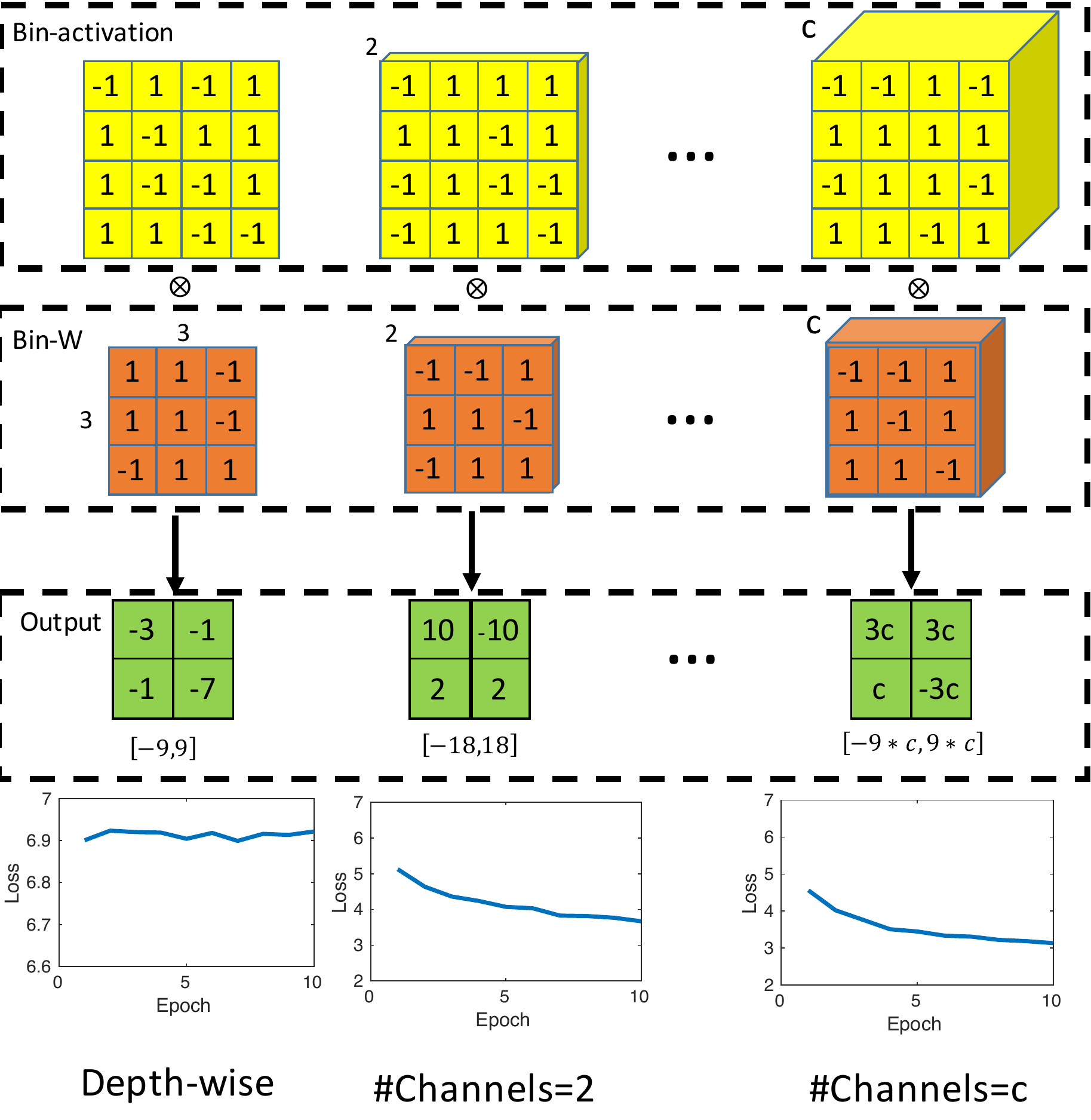}
    \caption{XNOR binary operation at depth-wise (left), group (middle), and fully (right) convolutional layers. $\bigotimes$ denotes the XNOR operator. Assuming all channels have the same value, the value range of the depth-wise convolution is constrained in $[-9,9]$, limiting the representation capability of BNNs but fast to convolve. The full binary convolution provides a larger possible value range, enhancing the capability of BNNs but slow to compute. The group convolution balances both worlds: it maintains a sufficiently wide value range to preserve the feature representation while being efficiently light-weight. The three figures in the bottom indicate the convergence behavior of the depth-wise, group and full convolution respectively during training phase. The binary depth-wise is prone to divergence; the full convolution effectively finds a way to the local mimima and is slightly better than the group convolution which steadily converges.}
    \label{fig:binaryConv}
\end{figure}
In the last few years, Deep Convolutional Neural Network (DCNN) for mobile platforms, which assumes certain constraints on computational capacity and battery, has been experimentally proven to be a successful approach in a wide variety of visual tasks in machine vision ~\cite{Iandola2017SqueezeNetAA,Howard-MobileNet, Sandler-MNetv2,mehta2018espnet,Zhang2018ShuffleNetAE, Ma_2018_ECCV}. Many compressed neural networks were proposed such as pruning ~\cite{Han-LBWs, Han-Deep, Liu2017LearningEC, Yu2019UniversallySN, Jiahui_ICLR_2019,liu2018rethinking} and quantization ~\cite{zhou2017, zhu2016trained, Li2017MimickingVE}. Binary Neural Networks (BNNs) recently have attracted many interests and achieved significant improvements ~\cite{Rastegari-xnor, Helwegen2019LatentWD, Liu2018BiRealNE,Adrian_bmvc_2019, Wang_2019_CVPR}. Prior works focused on binarizing large ConvNets which often contain several millions of parameters. On the other hand, compact neural network (e.g., MobileNet ~\cite{Howard-MobileNet}) is among promissing network architectures for binarization. The MobileNet exploits light-weight depth-wise and point-wise convolution layers to leverage the network efficiency when deploying on mobile devices. However, it is not trivial to make the depth-wise operators capable of coping with 1-bit quantization to push the network more compact.

With the depth-wise convolution, the neural network achieves low inference latency and even more optimal when being 1-bit quantized. However in such a binarization, input of the convolutional layers are channel-wise multiplied and summed. Therefore, the output values are limited within a narrow range. For instance, a binary $3\times3$ depth-wise filter convolving with one channel of the input yields values in $[-9,9]$, degrading the representation capability of the binary neural networks. On the other hand, the binary vanilla convolution results in a larger output value range which allows to attain an abundant feature representation and to effectively preserve the distribution of the data samples through network layers.  Figure~\ref{fig:binaryConv} illustrates the principle of this perspective. Although being effective, the vanilla convolution comes with an expensive computational cost since the filter convolves all channels of the input tensor. Therefore, the replacement of either depth-wise or vanilla by group convolution appears to be a promising approach to compensate the trade-off between the neural network latency, feature representation capability and computational resource constraint.

Group convolution is a simple yet efficient operation used in various neural networks to optimize trainable network parameters as well as the computation. AlexNet \cite{Alex_NIPS2012}, ConDenseNet \cite{Huang-ConDense}, ResNeXt \cite{Xie2016}, etc. are among popular neural architectures exploiting the group convolution and achieving great outcomes. At its extremity appears the depth-wise convolution. Inside the depth-wise, each channel is a separate group, or in other words the number of groups is exactly the depth of the input tensor. Vanilla convolution is also a special case of group convolution where $\mbox{\#groups}=1$. In most of the networks having group convolutional layers, the number of groups is often homogeneous at different layers in the network. From our perspective, a heterogeneous scheme to distribute groups at different layers can help to construct an efficient and accurate neural architecture, intuitively assuming a non-homogeneous feature representation through network layers. %

Aiming at leveraging the effectiveness of the heterogeneous group convolution, in this paper we propose a novel weight-sharing mechanism to explore in group search space optimally compact binary neural architectures that work efficiently and accurately. The key idea is to formulate the searching as an optimization problem that seeks to create a new genre of the compact architecture. This network is expected to be capable of performing efficiently in challenging and complicated tasks of image classification when data volume is huge and objects are diverse in types. Instead of conducting the search in a convolutional operation space with high degree of intractability as in neural architecture search (NAS) \cite{Barret_iclr17, pmlr-v80-pham18a, ying2019nasbench, Liu_PNAS_2018}, we exploit a controllable search space of group convolution in a MobileNet structure consisting of 13 layers \cite{Howard-MobileNet}, resulting in a potentially compact yet efficient binary architecture.

The main contributions of this paper are threefold:
\begin{itemize}
    \item We introduce a novel construction of binary neural network that is one of the first studies searching for a potential architecture design via a heterogeneous combination of group convolutional layers. Our work sheds the light on a new direction for enhancing the capability of BNNs.
    \item We propose an adaptive weight-sharing training mechanism that automatically searches in the group space to build efficient BNNs. More importantly, our training scheme is intuitive, flexible, and straightforward to implement. 
    \item We extensively conduct experiments to prove that following our approach, the binary neural architecture construction achieves a significant improvement factor regarding computation saving and model accuracy, therefore being able to attain state-of-the-art performance on large-scale ImageNet dataset \cite{imagenet_cvpr09}. 
\end{itemize}

\section{Related Work}
We have witnessed many research interests in binary neural networks. Courbariaux et al. \cite{CourbariauxBD15, Matt-BinaryNet, Itay-BNN} described the very first works to constrain full-precision weights in deep convolution neural networks to $\{-1,1\}$ by utilizing XNOR-count operator and being able to accelerate the inference stage $23\times$ faster than standard convolutional operation and $3.4\times$ than cuBlas~\cite{zhang2019dabnn}, an efficient GPU framework used for linear algebra computation. The work achieves high accuracy when benchmarking on popular datasets such as MNIST \cite{Lecun98gradient-basedlearning}, CIFAR10 \cite{Krizhevsky09}, SHVN \cite{Yuval-nips2011}. XNOR-Net \cite{Rastegari-xnor} is an interesting idea making use of scaling factors estimated from full-precision weights and achieving 44\% Top-1 accuracy on ImageNet with AlexNet architecture \cite{Alex_NIPS2012}. The two most related approaches to our works are Bi-RealNet \cite{Liu2018BiRealNE} and MoBiNet \cite{Hai_2020_WACV}. These binary models described a deployment of compact modules with skip connection and group convolution to enhance the capability of BNNs in terms of feature representation. The two models reach the state-of-the-art performance of 56\% and 54\% Top-1 accuracy on Imagenet respectively when binarizing both activations and weights. A recent work on BNNs \cite{Helwegen2019LatentWD} introduced Binary Optimizer to remove the dependency of binary weights from the real values, opening a new way to improve the BNNs.

To ameliorate the Binary Neural Network architecture, we adopt the methodology of Neural Architecture Search (NAS). The NAS aims at seeking to construct neural networks in an automatic instead of a manual manner. Many NAS algorithms formulate the search as an optimization problem and achieve great success \cite{liu2018darts, Chen2019pdarts} in finding  optimal architectures, assuming constraints on network latency and computational resource. In the following we focus on reviewing the neural architecture search appropriate to apply for mobile devices. Pham et al. introduced ENAS \cite{pmlr-v80-pham18a} considered as one of the first efficient neural architecture search approaches using cell-based search space. This network trains a super-graph from which sub-optimal paths are selected to create sharing parameters in sub-models. This mitigates the challenges when wandering in a huge exploration space by shrinking the search process parameters. There are other approaches outperforming manually designed networks. Liu et al. \cite{liu2018darts} proposed DARTS, a prominent gradient-based method that optimizes jointly one-shot models on a continuous relaxation of the search space. However because the models are assembled by a mixture from a set of operations, the performance relies heavily on the set selection. Another approach having the same flavor is ProxylessNAS \cite{cai2018proxylessnas} which adapts 1-bit neural architecture to abate GPU memory usage of one-shot models. The probability to select operation edges is updated by BinaryConnect \cite{CourbariauxBD15}.        

While NAS algorithms based on reinforcement learning and evolutionary methods strictly demand prohibitive computation with thousands of GPUs \cite{Barret_iclr17, Liu_PNAS_2018, Esteban_AAAI_2018, Real_ICML_EA_2017, MIIKKULAINEN2019293}, single-path one-shot architecture search methods are affordable over a conditional exploration space. Guo et al.~\cite{guo2019single} and Chen et al. \cite{Chen2019DetNASBS} implemented the one-shot model named SPOS and DetNAS to solve image classification and object detection problem, respectively. SPOS~\cite{guo2019single} delves into a random single path at every iteration to set up a super network on which the algorithm applies an evolutionary search to seek for an optimal path for neural network formation. In the one-shot network, pre-trained output can be used to transfer to different types of task like object detection and segmentation. Our proposed method has a similar flavor in training random group convolution, assuming modifications in the neural architecture with weight sharing and searching for the optimal group combinations.

\section{Our Methodology}
\subsection{Binary Operation}
In this section, we provide some fundamental background on binary neural network. When binarizing weights and activations, a typical binary neural network uses a sign function to constrain values to either $-1$ or $+1$.
\begin{equation} \label{eq:sign}
\xv^b = \text{Sign}(\xv) = \left\{\begin{matrix}
 +1,& & \text{if} \ \  \xv\geqslant 0, \\ 
 -1,& & \mbox{otherwise}
\end{matrix}\right.
\end{equation}
where $\xv^b$ is binarized value of $x$ which can be network inputs or weights. Similar to float-type neural network, 1-bit weights are intentionally computed to minimize an objective function:
\begin{equation}
    \wv^{b*} = \argmin_{\wv^b} \ L(f_b(\xv, \wv^b), y)
\end{equation}
where $L$ is the loss function; $\xv$, $\wv^b$, $y$ are inputs, binary weights, and labels respectively. Because 1-bit values degrade the neural network capability of preserving feature through layers, we apply scaling factors and backpropagation scheme mentioned in XNOR-Net \cite{Rastegari-xnor} to tackle the training divergence issue and to enhance the binary network performance. Also, to compute gradient of non-differentiable sign function, we adapt an approximation for the derivative of the sign function with respect to the activation \cite{Liu2018BiRealNE}.
\begin{equation}
\label{eq:update}
 \frac{\partial L}{\partial \Xv}=\frac{\partial L}{\partial \Xv^b}\frac{\partial \Xv^b}{\partial \Xv} = \frac{\partial L}{\partial \Xv^b}\frac{\partial \mbox{Sign}(\Xv)}{\partial \Xv} \approx  \frac{\partial L}{\partial \Xv^b}\frac{\partial A(\Xv)}{\partial \Xv}   
\end{equation}
$A(\cdot)$ denotes a differentiable approximation function in a piece-wise polynomial function~\cite{Liu2018BiRealNE}, where
\begin{equation} \label{eq:appro_sign}
\begin{split}
&A(x) = 
     \begin{cases}
       \text{-1,} &\quad\text{if x} < -1,\\
       \text{$2x+x^2$,} &\quad\text{if } -1 \le x < 0, \\
       \text{$2x-x^2$,} &\quad\text{if } 0 \le x < 1,\\
       \text{1,} &\quad\text{otherwise.} \\ 
     \end{cases}\\
&\frac{\partial A(x)}{\partial x} = 
    \begin{cases}
    \text{$2+2x$,} &\quad\text{if } -1 \le x < 0, \\
       \text{$2-2x$,} &\quad\text{if } 0 \le x < 1,\\
       \text{0,} &\quad\text{otherwise.} \\ 
    \end{cases}\\
\end{split}
\end{equation}
The weights are only binarized in forward step for both training and testing stage, then we can apply binary xnor-popcount operator \cite{MulaKL16, bmxnetv2} to accelerate the process. In backward step, real weights are stored to compute the derivatives and update new values.

\begin{figure}[t]
    \centering
    \includegraphics[scale=0.45]{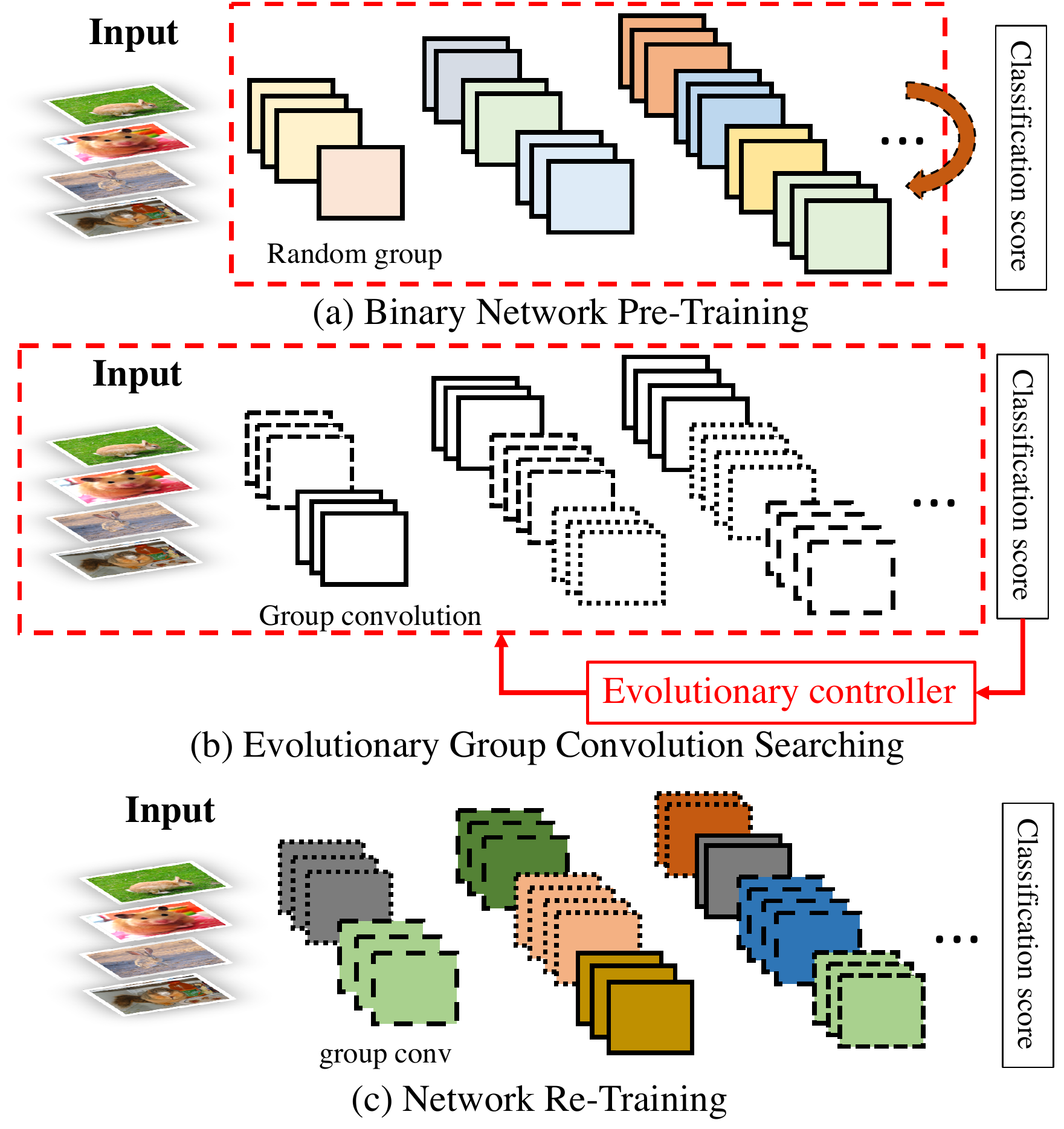}
    \caption{Framework overview of our proposed method. The process consists of three steps: train binary models with random groups (top figure, curve arrow indicates looping), apply evolutionary search to explore optimal groups based on accuracy metric, and re-train the searched group models from scratch.}
    \label{fig:bin_framework}
\end{figure}

\subsection{Design And Search 1-Bit MobileNets}
Binary neural network and neural architecture search are two among the most potential techniques used to construct compact yet efficient neural models. Network architecture design usually has a great impact on the performance of the binary networks. The main objective in our work is to explore efficient designs of BNNs with the hope that techniques in neural architecture search (NAS) can leverage the exploration for compact structures. However, the NAS often covers a huge search space of convolutional operators so that it is able to generate sub-optimal neural networks. This might be very difficult and costly when directly exploring in the binary operator space. To simplify the search space and to prevent the computation from exorbitant price, we develop a novel training procedure exploiting randomized group convolution operators with weight sharing in the neural networks. With this approach, the binary models become robust thanks to the consideration of a wide variety of group combinations which fosters the group search procedure. Exploring new architecture for binary neural networks using neural architecture search can open a potential research direction to significantly improve the binary network construction. In the next sections, we discuss how to conduct and optimize the group convolution search with our proposed training pipeline.
\subsubsection{Evolutionary Group Convolution Search for Binary Neural Networks}
To our knowledge, evolutionary algorithms, a.k.a genetic algorithms, base on the well-known evolution of creature species in nature. Natural selection eliminates individuals unable to adapt to the environment. Additionally, survivals are kept for reproduction, crossover, and mutation.
Several recent evolutionary approaches for neural architecture are proposed \cite{liu2019metapruning, Xie_ICCV_EA_2017}. Instead of searching for the entire network including a complete set of connections and operators as in prior works, we conduct an evolutionary search for group values at convolutional layers to explore suitable binary structure with a simple and effective network design. At each layer, group candidatures consist of all possible divisors of the input channels. In detail, we start by sampling a list of possible groups and searching on this list to find an optimal architecture by training random groups for every iteration. The first objective is to achieve an accuracy superior than a threshold. Second, in order to make the computational cost controllable, we select binary compact models satisfying certain constraint on the maximum number of FLOPs such that
\begin{equation}
\mbox{FLOP}(\Wv, \mathcal{G}) \leq \mbox{FLOP}_{\mbox{max}}
\end{equation}
where $\Wv$ is weight and $\mathcal{G}$ is a group combination for each convolutional layer respectively. The search pipeline is presented in Algorithm~\ref{code:search}. 
\begin{algorithm}
  \caption{Evolutionary Search for Group Convolution}
  \label{code:search}
  \textbf{Input}: Candidate Group Size: $\mathcal{S}$, Top Candidates: $\mathcal{K}$, \#Crossovers: $\mathcal{C}$, \#Mutations: $\mathcal{M}$, Model weights: $\mathcal{W}$, and FLOP constraint: $\mathcal{F}$ \\
\textbf{Output}: Optimal group combination $\mathcal{G}^*$ that yields top accuracy among the other combinations.
  \begin{algorithmic}[1]
    \STATE $\mathcal{G}^*  \leftarrow  \textit{Sample\_Candidates}(\mathcal{S}, \mathcal{F})$ \ \ \# group candidates
    \FOR{i=1:maxIteration}
    \STATE Fitness $\leftarrow \textit{Accuracy}(\mathcal{W}, \mathcal{G}^*)$ \# accuracies 
    \STATE $\mathbf{K} \leftarrow \textit{Select\_TopK}(\mbox{Fittness}, \mathcal{K})$
    \STATE $\mathbf{C} \leftarrow \textit{Crossover}(\mathbf{K}, \mathcal{C})$
    \STATE $\mathbf{M} \leftarrow \textit{Mutation}(\mathbf{C}, \mathcal{M})$
    \STATE $\mathcal{G}^* \leftarrow \mathbf{C} \cup \mathbf{M}$
    \ENDFOR
  \end{algorithmic}
\end{algorithm}

\subsubsection{Module Modification}
\label{me:module}
\begin{figure}[t]
    \centering
    \includegraphics[scale=0.36]{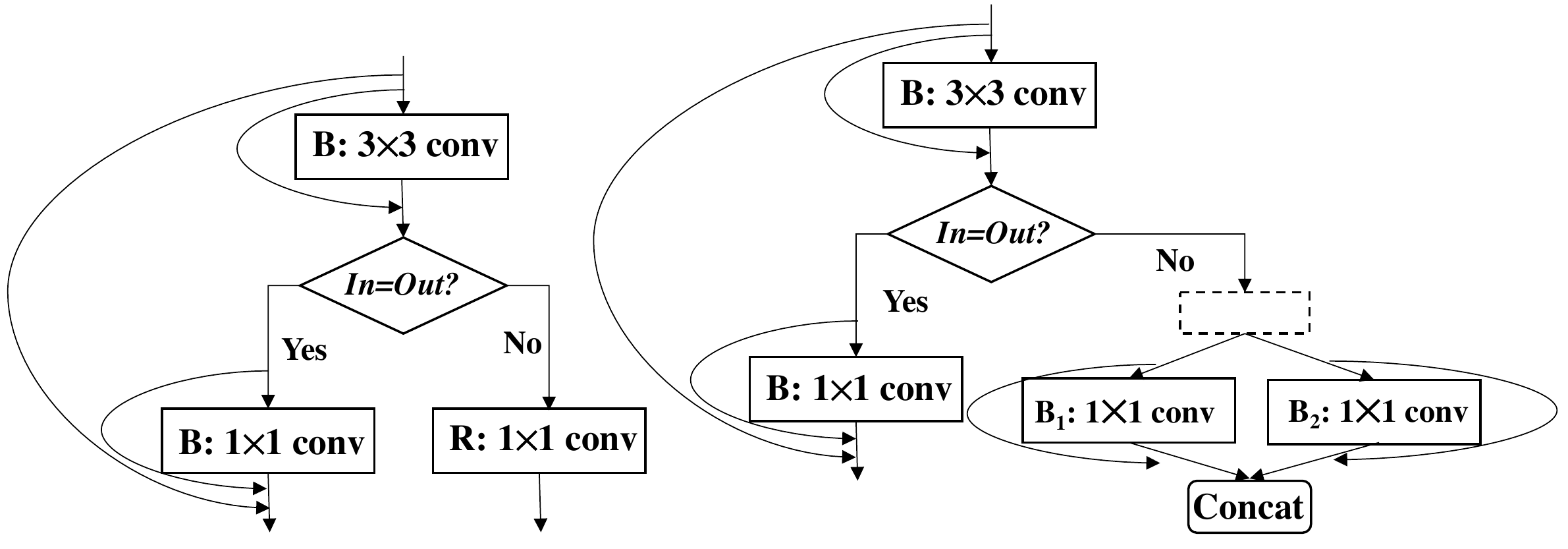}
    \caption{Illustration of network module modification.}
    \label{fig:module}
\end{figure}

MobileNets \cite{Howard-MobileNet} with depth-wise and point-wise convolution (together known as separable depth-wise convolution \cite{Sifre}) are famous for its compactness and effectiveness when being used for designing a neural network. We modify the MobileNet structure to facilitate the creation of efficient binary neural networks that outperform prior state-of-the-art works regarding accuracy and memory saving. However, training a binary depth-wise convolution is not straightforward \cite{Hai_2020_WACV} because the separate channel-wise output falls into a small value range due to the nature of the computation, making the binary network impossible to converge. To overcome this issue, we propose a replacement of depth-wise convolution by group convolution to enlarge the value range of the depth-wise convolution output. More precisely, we search for groups of binary convolution operators of kernel size $3\times3$ and $1\times1$. To preserve the feature representation through binary layers while assuming a low computational cost, we maintain the full precision $1\times1$ convolution when perceiving a reduction in spatial dimension at a layer output. In addition, block-wise and layer-wise skip connections are added in case of homogeneous dimension output to benefit the network training. Our proposed network module is illustrated in Figure~\ref{fig:module}. There are three principle modifications that make our modules different from the vanilla architecture of the MobileNet \cite{Howard-MobileNet}:
\begin{itemize}
    \item \textbf{Module 1 (M1)}: consists of a binary $3\times3$ group conv and a binary $1\times1$ full conv. A real $1\times1$ fully conv follows when there is a spatial dimension shrinkage. (see Figure~\ref{fig:module} - the left figure).
    \item \textbf{Module 2 (M2)}: uses group convolution for real $1\times1$ full conv to further reduce the computational cost. The group is also searched along with the binary $3\times3$ group conv. 
    \item \textbf{Module 3 (M3)}: is made up of two binary $1\times1$ conv layers instead of one, and then concatenate them to obtain the same dimension.
\end{itemize}
In the next section, we describe a training scheme based on randomized group through weight sharing to force the binary neural network to converge.
\subsubsection{Randomized Groups via Weight-sharing}
\begin{figure}
\begin{center}
\subfigure{
    \hspace{-0.5em}\includegraphics[scale=0.38]{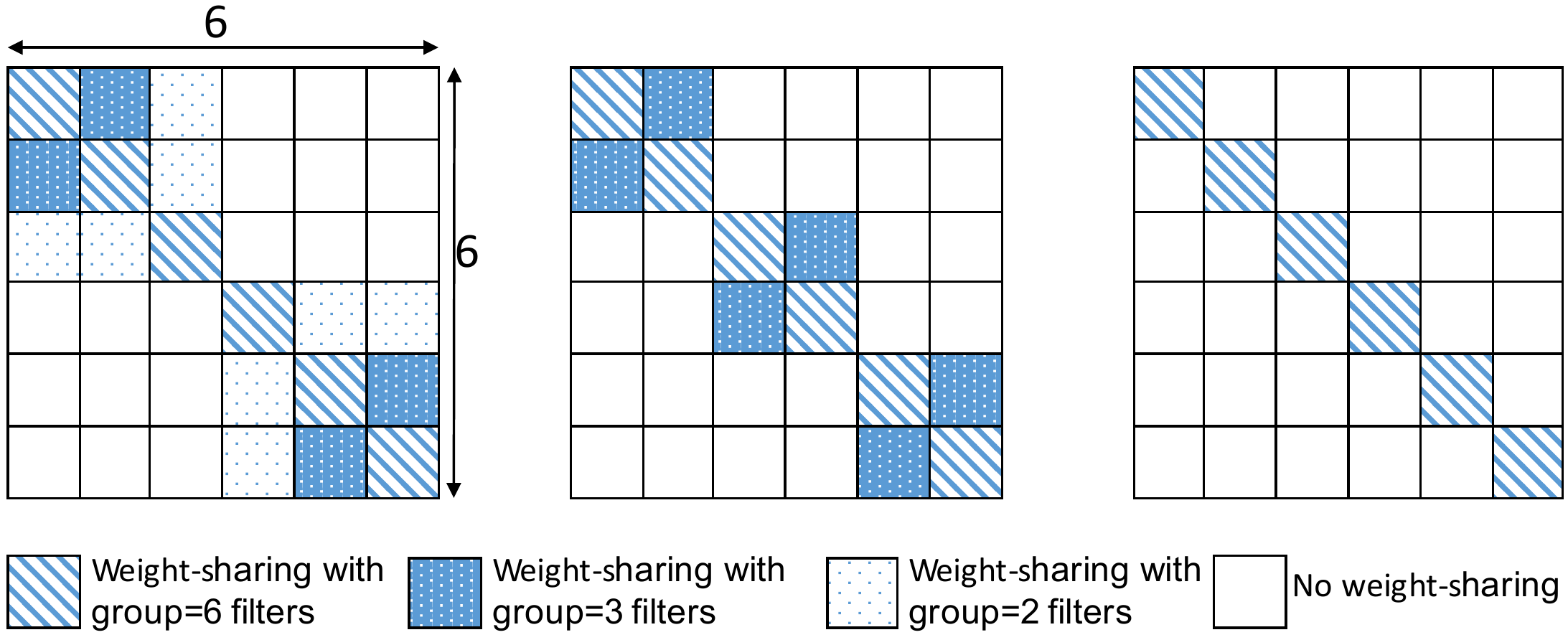}
}\hspace{0.5cm}
\end{center}
    \caption{Illustration of a 2D weight sharing. For example, in a $6\times6$ filter, weights can be shared within groups of $2$, $3$, and $6$.}
    \label{fig:weight-share}
\end{figure}

In the search stage, a fitness function (e.g., accuracy of the model) is computed to help explore optimal group combinations. However if we naively calculate accuracy of a binary model without training with data samples (i.e., images), it does not guarantee that the optimal model is able to learn the distribution of the target dataset. Therefore, to leverage important information from a given dataset for evolutionary search, we propose a method to train the binary model along with randomized group combination via weight-sharing in each training iteration. To ease the implementation, full convolutions are initialized and cropped with randomized groups in each iteration via weight-sharing. The weight-sharing is depicted in Figure~\ref{fig:weight-share}.   
\subsubsection{Training Procedure}
Binary neural network is an active and progressive research topic with prominent works \cite{Hai_2020_WACV, Liu2018BiRealNE, Wang_2019_CVPR}. Training neural networks of 1-bit weights is a difficult task because feature representation often has narrow value range which seems to be impossible to fit the target large-scale dataset for classification. XNOR-Net~\cite{Rastegari-xnor} utilizes full precision weight values to derive real value scaling factors which play crucial role in amplifying the magnitude of binary weights and activations. The optimization is formulated as follows:
\begin{equation}
\label{eq:xnor_optim}
    \Xv^{b*}, \alphav = \argmin_{\Xv^b \in \left \{-1,1  \right \} , \alphav > 0}\left \| \Xv-\alphav \Xv^b \right \|_2^2
\end{equation}
$\Xv$ can be weights or activations and $\alphav$ are scaling factors. The optimal solution for Equation \ref{eq:xnor_optim} is $\Xv^{b*}=\mbox{Sign}(\Xv)$ and $\alphav =\frac{1}{(\Xv^b)^{T}\Xv^b}\left | \Xv \right |_{l1}$. Bi-RealNet \cite{Liu2018BiRealNE} and MoBiNet~\cite{Hai_2020_WACV} make use of skip connections to enhance the performance of binary neural networks. With that flavor, we manipulate the skip connections together with scaling factors to facilitate the training procedure. Here follows the summary of our training: 
\begin{itemize}
    \item For each iteration, we train binary neural networks with random group combinations. For instance, if the network has 13 layers, the groups corresponding to these layers are randomized within possible divisors of input channels. This randomization helps 1-bit models become robust against group changes when searching.
    \item Evolutionary search described in Algorithm \ref{code:search} is applied to seek for optimal groups. An ablation study is conducted in Section~\ref{exp:ablation} to prove that our search approach is more efficient than arbitrary randomized groups.
    \item We train from scratch the final binary models with optimal group convolution. All steps run on large scale ImageNet-1k \cite{imagenet_cvpr09}.
\end{itemize}
The training pseudo-code is illustrated in Algorithm \ref{code:training} and visualized in Figure \ref{fig:bin_framework}.
\begin{algorithm}
  \caption{Overall Training BNNs}
  \label{code:training}
  \textbf{Input}: Full binary neural model and inputs for evolutionary search   \\
\textbf{Output}: New optimal binary neural model with new group structure.
  \begin{algorithmic}[1]
    \STATE $\textit{C\_ins} \leftarrow \textit{Input\_Channels}$
    \STATE Initialize Binary Models  $\mathcal{M}_b$
    \FOR{i=1:Iteration}
    \STATE $\mathcal{G}_r \leftarrow \textit{Random\_Groups}(\textit{C\_ins})$  \ \ \# random group
    \STATE $\textit{Train\_Group}(\mathcal{M}_b, \mathcal{G}_r)$ 
    \ENDFOR
    \STATE $\mathcal{G}^* \leftarrow \textit{Search\_Group}(\mathcal{M}_b)$   \ \ \# search group using Algorithm \ref{code:search}
    \STATE $\textit{Train\_Group}(\mathcal{M}_b, \mathcal{G}^*)$ 
  \end{algorithmic}
\end{algorithm}
\section{Experiments}
In this section, we demonstrate the performance evaluation of our proposed method. First, we describe experiment setups and implementation details. Second, to prove our weight-sharing group search mechanism more effective and reasonable than naively random search we compare the training performance with randomized groups in ablation studies. Third, we evaluate the search groups with uniform normal groups to investigate the fact that for each level of feature representation, the number of groups should be different. Then we compare with the state-of-the-arts to see improvement impacts of our proposed BNNs. Finally, the computation analysis are presented. All experiments including searching, training, and testing are conducted on the large-scale dataset of ImageNet2012-1k. We analyzed the results regarding three metrics: Top-1, Top-5 classification accuracy on ImageNet dataset, and number of FLOPs.            
\subsection{Experimental Setups}
\label{exp:setups}
\noindent\textbf{Dataset}. The image dataset we used to demonstrate the effectiveness of our framework is ILSVRC2012 \cite{imagenet_cvpr09}, a dataset containing 1.2M and 50K image samples for training and testing respectively. The dataset has 1000 classes. Most of the previous works such as XNOR-Net~\cite{Rastegari-xnor}, Bi-RealNet~\cite{Liu2018BiRealNE}, CI-BCNN~\cite{Wang_2019_CVPR}, and MoBiNet~\cite{Hai_2020_WACV} also used this dataset to evaluate their model performance.\\

\noindent\textbf{Implementation details and setups}. Our training pipeline consists of three main stages: train binary architecture with randomized groups, search groups for convolutional layers via evolutionary method, and train the final models with searched groups from scratch. We train on basic blocks modified from MobileNet to improve the performance of binary models, mentioned in Section \ref{me:module}. Each image is scaled up to $256\times256$. In training, images are randomly cropped to $224\times224$. In testing, they are centrally cropped to $224\times224$. When training, real-valued filters are saved in RAM to compute update values in backpropagation via Equation~\ref{eq:update} and then are binarized in inference stage. In the first stage of the training pipeline, we used batch size of $512$ images to train the 1-bit models with random groups and learning rate of $0.001$ in $64$ epochs. In the search stage, the number of populations, crossovers, mutations are $50$, $25$, $25$ respectively and the searching runs $20$ iterations to find optimal group structure. In the end of the pipeline, we train the final models from scratch with batch size $512$, learning rate $0.001$, number of epochs $256$. All training stages use Adam Optimizer~\cite{kingma2014method},  momentum $0.9$, and update learning rate through linear decay. FLOPs are calculated following the suggestion of \cite{Liu2018BiRealNE, Wang_2019_CVPR} for fair comparison. Training is conducted on four RTX 2080 Ti GPUs 24GB and searching is on one GPU.  

\begin{figure}[t]
    \centering
    \includegraphics[scale=0.5]{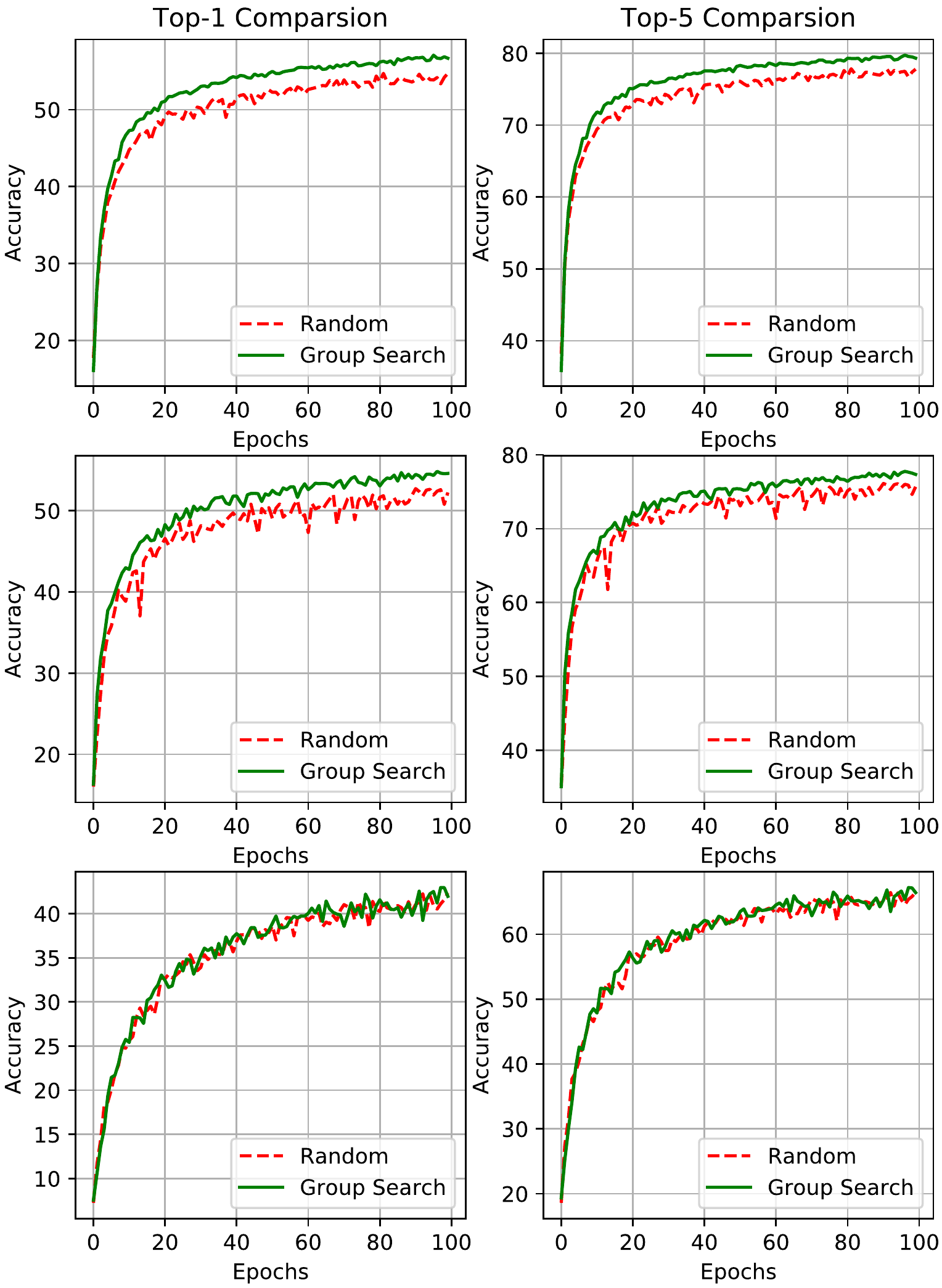}
    \caption{Validation accuracy on ImageNet through epochs of modified modules (M1, M2 and M3, from top to bottom respectively). Random: Groups are naively randomized for layers. Group search: Our proposed search optimal group architecture.}
    \label{fig:exp1_compare_rand}
\vspace{-1em}
\end{figure}
\subsection{Our Search Group vs. Random Group Search and Uniform Group Architectures}
\label{exp:ablation}
To investigate our hypothesis of binarizing convolution via an evolution-based searching in MobileNet's architecture, we compare with random search and uniform group architectures as an ablation study. The experiment is conducted on the three proposed modules in Section \ref{me:module}. For the Module 2, there are four full-precision $1\times1$ convolutions. We also apply searched groups for such layers to further reduce the computational cost. 

To compare with random group search, we report Top-1 and Top-5 accuracy for each epoch. The training performance comparison is indicated in Figure \ref{fig:exp1_compare_rand}, showing the result of modifications from MobileNets: Module 1, Module 2, and Module 3 (from the top to the bottom in that order). 

We run the comparison experiments of randomized and searched groups with $100$ epochs, $512$ for batch size and observe the Top-1 and Top-5 accuracy in ImageNet validation set for each epoch. With respect to the first two modules, our group search architecture training is more stable and for all epochs, we achieve more accurate results (about $2\%$) in both Top-1 and Top-5 accuracy.

Our proposed search group achieves better performance when comparing with random groups that require more computational cost. Table~\ref{tab:mem_flops_compa} reports the number of FLOPs when running with random group and with our proposed group search. Regardless of the fact that random architectures have a larger computational cost, our search group networks are more accurate and efficient. 

We also provide a comparison with uniform group (i.e., using the same number groups for all layers of Module M1, M2, and M3) as a ablation study for investigating our hypothesis. We train models with uniform groups of $1$ (fully convolution), $4$, and $16$. The Table \ref{tab:mem_flops_compa} presents the results of Top-1, top-5 accuracy, and number of FLOPs (computational cost). 

Our reported statistics expresses a trade-off of performance between fully convolution and depth-wise convolution. For example, in M1 and M3 our searched group models outperform comparable uniform group models (g=4 and 16) in accuracy and take less FLOPs.   

\begin{table}[t]
\hspace*{-1.0cm} 
\vspace{-1em}
\begin{center}
\begin{tabular}{|l|c|c|c|}
\hline
\#Groups (M1) & Top-1 (\%) & Top-5 (\%) & FLOPs\\
\hline\hline
Groups = 1 & 64.51 & 85.14 & $2.13 \times 10^8$\\
Groups = 4 & 60.89 & 82.54 & $1.63 \times 10^8$\\
Groups = 16 & 58.49 & 80.66 & $1.50 \times 10^8$\\
Random Group & 59.05& 81.22& $1.58 \times 10^8$\\
\textbf{Ours} & \textbf{60.90} & \textbf{82.60} & \textbf{1.54$\times 10^8$}\\ 
\hline\hline
\#Groups (M2) & Top-1 (\%) & Top-5 (\%) & FLOPs\\
\hline\hline
Groups = 1 & 64.51 & 85.14 & $2.13 \times 10^8$\\
Groups = 4 & 59.59 & 81.67 & $0.67 \times 10^8$\\
Groups = 16 & 54.23 & 77.04 & $0.30 \times 10^8$\\
Random Group & 58.13 & 80.42& $0.75 \times 10^8$\\
\textbf{Ours} & \textbf{59.30} & \textbf{81.00} & \textbf{0.62$\times 10^8$}\\
\hline\hline
\#Groups (M3) & Top-1 (\%) & Top-5 (\%) & FLOPs\\
\hline\hline
Groups = 1 & 57.56 & 79.85 & $0.87 \times 10^8$ \\
Groups = 4 & 49.90 & 73.15 & $0.37 \times 10^8$ \\
Groups = 16 & 45.29 & 69.38 & $0.24 \times 10^8$ \\
Random Group & 50.07 & 74.11 & $0.38 \times 10^8$ \\
\textbf{Ours} & \textbf{51.06} & \textbf{74.18} & \textbf{0.33$\times 10^8$}\\
\hline
\end{tabular}
\end{center}
\bigskip
\vspace{-1em}
\caption{Uniform grouping baselines and random group search vs Our group search on Module M1,M2, and M3.}
\vspace{-1em}
\label{tab:mem_flops_compa}
\end{table}

\subsection{The Efficiency of Our BNNs}
MobileNet architecture is a compact network working accurately and efficiently based on light-weight module of separable convolution layers. Binarizing such a compact model can give us promising outcomes because it contains less parameters thanks to the tremendous reduction of the computational cost without incurring accuracy loss. However, as mentioned in Section~\ref{sec:intro} the networks exploiting the separable convolutions including depth-wise scheme cannot convergence when being binarized because of extremely small value range that cannot adequately fit complex data samples like images. On the contrary, groups and fully convolutional layers are easier to make the networks perform well. Albeit achieving high accuracy, fully convolutional layers are not efficient to deploy on mobile devices because of a huge number of parameters. So, group convolutional layers can have potential trade-off between depth-wise and full convolution. In this work, we propose a group search mechanism via evolutionary method to find group structure at each convolutional layers for a binary neural network in the MobileNet architecture.

For showing the effectiveness of our proposed search mechanism, we conduct experiments of modified modules with different computational budget constraints. We firstly train binary models with random groups for each module in $64$ epochs. Then, we search for networks satisfying the FLOP budget to derive optimal group structures. Finally, the networks with optimal groups are trained from scratch in $256$ epochs. The other settings are mentioned in Section \ref{exp:setups}. Top-1, Top-5 accuracy on ImageNet-1k, FLOPs, budget constraint, and number of GPU-hours of searching are reported in the Table \ref{tab:img_re}.

Compared to the full-precision MobileNet \cite{Howard-MobileNet}, our constructed binary neural networks accelerate approximately $4\times$, $9\times$, and $17 \times$ when using Module 1, Module 2, and Module 3 respectively, while incurring small Top1-accuracy loss of $10\%$, $11.6\%$, and $19.8\%$. We also outperform the most related work of MoBiNet \cite{Hai_2020_WACV}. This detail is mentioned in Section \ref{exp:comp_state}.  In addition, our search algorithm only takes $\approx$ 29h on one GPU in average.

In the next section, we compare our method with other state-of-the-art binary neural networks.
\begin{table}[t]
\centering
\begin{tabular}{ |l|c|c|c| } 
\hline
M & M1 & M2 & M3 \\
\hline
Top-1 (\%)& 60.9 & 59.3 & 51.1\\ 
Top-5 (\%)& 82.6 & 81.0 & 74.2  \\
FLOPs & $1.54 \times 10^8$ & $0.62\times10^8$ & $0.33\times10^8$  \\
MaxFLOPs & $1.55\times10^8$ & $0.80\times10^8$ & $0.50\times10^8$ \\
\#GPU-hours & 30 & 32 & 26  \\
\hline
\end{tabular}
\bigskip
\caption{The efficiency of proposed module M1, M2 and M3 in searched group architecture. The results are conducted on large scale of ImageNet dataset. MaxFlops is the constraint budget.}
\label{tab:img_re}
\vspace{-1em}
\end{table}
\subsection{Comparison with State-of-the-art Methods}
\label{exp:comp_state}
Binary neural networks make an amazing progress when recently achieving impressive results. However, prior works improve binary models through training process for representation learning while the architecture design should has great influence as well. Our proposed method using evolutionary search based on recent ideas of one-shot neural architecture search aims at exploring the group architecture design for BNNs improvement.

In this section, to evaluate the proposed method we compare our BNNs with several recent works: Binary Connect \cite{CourbariauxBD15}, BNNs \cite{NIPS2016_6573_BNN}, ABC-Net \cite{NIPS2017_6638_TowardsAcc}, DoReFa-Net \cite{zhou2016dorefa},  XNOR-Net \cite{Rastegari-xnor}, etc. The metrics reported are Top-1, Top-5 accuracy on ImageNet, and the number of FLOPs. BiReal-Net and CI-BCNN are two prominent works achieving good results. These networks binarize ResNet \cite{Kaiming_ResNet} with efficient skip connection module. Here, we only consider ResNet 18 layers versus our 13 layers for fair comparison. CI-BCNN \cite{Wang_2019_CVPR} is the state-of-the-art binary model (both weights and activations are binarized) as it is able to achieve $59.90\%$ Top-1 accuracy on ImageNet with very small cost of $1.54\times10^8$ FLOPs. Our binary model using Module 1 outperforms the MoBiNet \cite{Hai_2020_WACV} $6\%$ and  the Bi-RealNet-18 \cite{Liu2018BiRealNE} $4\%$ Top-1 accuracy with less computational cost. Moreover, it also surpasses CI-BCNN \cite{Wang_2019_CVPR} $1\%$ Top-1 accuracy with lower number of FLOPs (ours: $1.54\times10^8$, CI-BCNN: $>1.54\times10^8$). Also, our Module 2 and Module 3 also transcends the BiReal-Net \cite{Liu2018BiRealNE} by requiring a significant lower number of FLOPs.  

On the other hand, our method significantly outperforms the other binary neural networks regarding the accuracy and computation metric. The accuracy results are reported in the Table \ref{ta:com_state}. Our proposed binary networks are better than most of the prior works. For computational cost, Table \ref{tab:mem_flops_compa} indicates comparisons in terms of number of FLOPs and memory usage.  
\begin{table}[t]
\vspace{-1em}
\hspace*{-1.0cm} 
\begin{center}
\begin{tabular}{|l|c|c|c|}
\hline
\specialcell{Networks} & W/A &Top-1&Top-5\\
\hline\hline
Binary Connect~\cite{CourbariauxBD15}& 1/32&35.40&61.00\\
BWN~\cite{Rastegari-xnor}&1/32&56.80&79.40\\
BNNs~\cite{NIPS2016_6573_BNN}&1/1& 42.20&67.10\\
ABC-Net~\cite{NIPS2017_6638_TowardsAcc}&1/1&42.70&67.60\\
DoReFa-Net~\cite{zhou2016dorefa}&1/1&43.60&-\\
SQ-BWN~\cite{Dong2019}&1/1&45.50&70.60\\
XNOR-AlexNet~\cite{Rastegari-xnor}&1/1&44.20 &69.20\\
XNOR-ResNet-18~\cite{Rastegari-xnor}&1/1& 51.20&73.20\\
MoBiNet \cite{Hai_2020_WACV} &1/1&54.40&77.50\\
Bi-RealNet-18~\cite{Liu2018BiRealNE}&1/1&56.40&79.50\\
CI-BCNN-18 ~\cite{Wang_2019_CVPR}&1/1&56.73&80.12\\
CI-BCNN-18 (add) ~\cite{Wang_2019_CVPR}&1/1&59.90&84.18\\
\hline\hline
\specialcell{\textbf{Ours (M3)}} &1/1&\textbf{51.06}&\textbf{74.18}\\
\specialcell{\textbf{Ours (M2)}} &1/1&\textbf{59.30}&\textbf{81.00}\\
\specialcell{\textbf{Ours (M1)}} &1/1&\textbf{60.90}&\textbf{82.60}\\
\hline
\end{tabular}
\end{center}
\bigskip
\vspace{-1em}
\caption{The Top-1 and Top-5 accuracy comparison between the state-of-the-art and our method. Our Module 1 (M1) outperforms the state-of-the-art CI-BCNN \cite{Wang_2019_CVPR} by $1\%$ Top-1 accuracy.}
\vspace{-1em}
\label{ta:com_state}
\end{table}

\begin{table}[t]
\hspace*{-1.0cm} 
\begin{center}
\begin{tabular}{|l|c|c|}
\hline
\specialcell{Networks}& \specialcell{FLOPs}\\
\hline\hline
XNOR-AlexNet~\cite{Rastegari-xnor}&$1.38\times10^8$\\
XNOR-ResNet-18~\cite{Rastegari-xnor}&$1.67\times10^8$\\
Bi-RealNet-18~\cite{Liu2018BiRealNE}&$1.63\times10^8$\\
CI-BCNN-18 ~\cite{Wang_2019_CVPR}&$1.54\times10^8$\\
CI-BCNN-18 (add) ~\cite{Wang_2019_CVPR}&$> 1.54\times10^8$\\
MoBiNet \cite{Hai_2020_WACV}&$0.52\times10^8$\\
\hline\hline
\specialcell{\textbf{Ours (M3)}}&$0.33\times10^8$\\
\specialcell{\textbf{Ours (M2)}}&$0.62\times10^8$\\
\specialcell{\textbf{Ours (M1)}}&$1.54\times10^8$\\
\hline
\end{tabular}
\end{center}
\bigskip
\vspace{-1em}
\caption{Computational cost comparison between the state-of-the-arts and our method.}
\vspace{-1em}
\label{ta:com_compu}
\end{table}
\subsection{Analysis}
\label{exp:analy}
In this section, we discuss the analysis of results and computational complexity. For ablation study in Section~\ref{exp:ablation}, the results of group search architecture are more stable and have higher accuracy than naively erratic groups, proving that having heterogeneous group structure at each layer in MobileNet architecture yields good performance. In addition, group convolution is flexible to increase or decrease the number of connections in selective layers. For example when observing the first layers, we realize that the search algorithm tends to assign small number of groups to preserve essential information of the inputs. Meanwhile, the algorithm diminishes insignificant inter-channel connections (i.e., by increasing the number of groups) to enhance the model's compactness and efficiency.

From Table~\ref{tab:mem_flops_compa} and Table~$\ref{ta:com_state}$, our module 2 and module 3 have higher accuracy than the two prominent works of BiReal-Net \cite{Liu2018BiRealNE} and CI-BCNN \cite{Wang_2019_CVPR}. Moreover, the modules have a much lower computational cost ($\approx$ 33M FLOPs), approximately $5\times$ speed up factor when comparing with BiReal-Net (163M FLOPs).

\section{Conclusion}
Efficient group design for BNNs can yield good outcomes. We introduced a novel algorithm via evolutionary search to explore group structures aiming at optimizing the trade-off when either using depth-wise or fully convolutional layers in MobileNet. Our BNN is efficient as it achieves highly accurate results while saving the computational cost (only single GPUs for searching) in dealing with challenging visual classification tasks. \\ 
\textbf{Acknowledgements}. We thanks all anonymous reviewers for constructive and valuable feedback. The code will be available at \href{https://github.com/HaiPhan1991/BinMobileNet_Evo_Search}{link}

{\small
\bibliographystyle{ieee_fullname}
\bibliography{egpaper_final}
}

\end{document}